\documentclass[sigconf]{acmart}
\usepackage{multirow}
\usepackage{graphicx}
\usepackage{float}
\usepackage{subfig}
\usepackage{blindtext}
\usepackage{booktabs,capt-of}
\AtBeginDocument{%
  }

\copyrightyear{2025}
\acmYear{2025}
\setcopyright{acmlicensed}
\acmConference[MM '25]{Proceedings of the 33rd ACM International Conference on Multimedia}{October 27--31, 2025}{Dublin,Ireland}
\acmBooktitle{Proceedings of the 33rd ACM International Conference on Multimedia (MM '25), October 27--31, 2025, Dublin, Ireland}
\acmDOI{10.1145/3746027.3754727}
\acmISBN{979-8-4007-2035-2/2025/10}

\begin{document}

\title{Chain-of-Cooking: Cooking Process Visualization via Bidirectional Chain-of-Thought Guidance}

%\email{webmaster@marysville-ohio.com}
%\affiliation{%
%  \institution{Institute for Clarity in Documentation}
%  \country{USA}
%}

\author{Mengling Xu}
\orcid{0009-0002-4809-9129}
\affiliation{%
  \institution{Nanjing University of Posts and Telecommunications}
  % \streetaddress{1 Th{\o}rv{\"a}ld Circle}
  \city{Nanjing}
  \country{China}}
\email{2022010217@njupt.edu.cn}

\author{Ming Tao}
\orcid{0000-0002-4662-7170}
\affiliation{%
  \institution{Nanjing University of Posts and Telecommunications}
  % \streetaddress{1 Th{\o}rv{\"a}ld Circle}
  \city{Nanjing}
  \country{China}}
\affiliation{
  \institution{Peng Cheng Laboratory}
  \city{Shenzhen}
  \country{China}}
\email{mingtao2000@126.com}

\author{Bing-Kun Bao}
\authornote{Corresponding author}
\orcid{0000-0001-5956-831X}
\affiliation{%
  \institution{Nanjing University of Posts and Telecommunications}
  \city{Nanjing}
  \country{China}}
\affiliation{
  \institution{Peng Cheng Laboratory}
  \city{Shenzhen}
  \country{China}
  }
\email{bingkunbao@njupt.edu.cn}

\renewcommand{\shortauthors}{Xu et al.}

\begin{abstract}
Cooking process visualization is a promising task in the intersection of image generation and food analysis, which aims to generate an image for each cooking step of a recipe. However, most existing works focus on generating images of finished foods based on the given recipes, and face two challenges to visualize the cooking process.  First, the appearance of ingredients changes variously across cooking steps, it is difficult to generate the correct appearances of foods that match the textual description, leading to semantic inconsistency. Second, the current step might depend on the operations of previous step, it is crucial to maintain the contextual coherence of images in sequential order. In this work, we present a cooking process visualization model, called \textbf{Chain-of-Cooking}. Specifically, to generate correct appearances of ingredients, we present a Dynamic Patch Selection Module to retrieve previously generated image patches as references, which are most related to current textual contents. Furthermore, to enhance the coherence and keep the rational order of generated images, we propose a Semantic Evolution Module and a Bidirectional Chain-of-Thought (CoT) Guidance. To better utilize the semantics of previous texts, the Semantic Evolution Module establishes the semantical association between latent prompts and current cooking step, and merges it with the latent features. Then the CoT Guidance updates the merged features to guide the current cooking step remain coherent with the previous step. Moreover, we construct a dataset named \textbf{CookViz}, consisting of intermediate image-text pairs for the cooking process. Quantitative and qualitative experiments show that our method outperforms existing methods in generating coherent and semantic consistent cooking process. 
\end{abstract}



\begin{teaserfigure}
   \includegraphics[width=\textwidth]{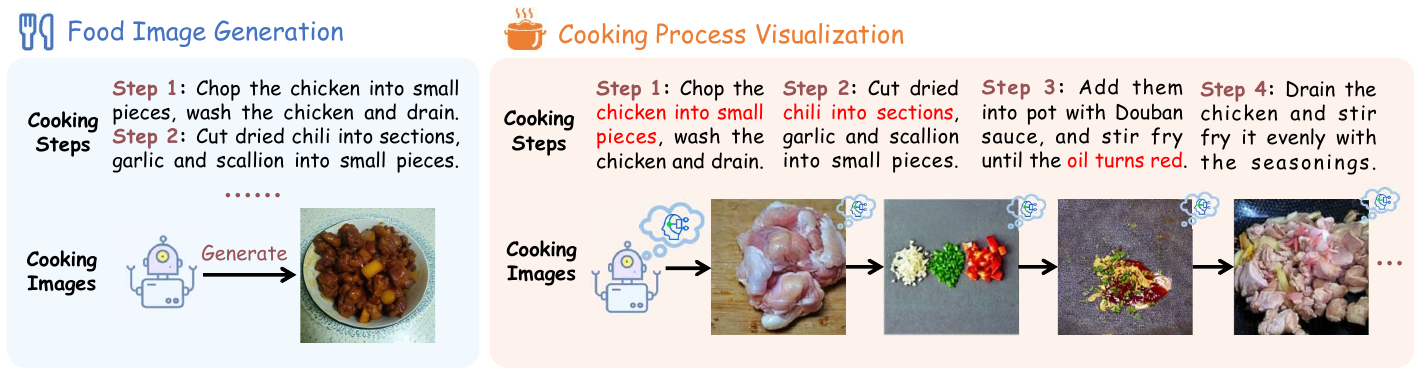}
   \caption{Tasks comparison of traditional food image generation (left) and the cooking process visualization (right).}
   \label{fig1}
\end{teaserfigure}


\maketitle

\section{Introduction}
Cooking process visualization has emerged as a novel derivative task in the domain of food analysis, garnering increasing attention in recent years. As shown in Figure \ref{fig1}, given the cooking process of a certain food, the cooking process visualization model outputs a sequence consisting of images corresponding to each step of the operation. This capability holds significant promise for cooking education and dietary planning, as it helps users follow complex cooking instructions more intuitively. 

Numerous studies have been conducted on food image generation \cite{foodfusion,clusdiff,cookgalip, ml-cookgan} and witnessed remarkable progress in the fidelity and realism of generated images \cite{xu2025sd}. For instance, FoodFusion \cite{foodfusion} and ClusDiff \cite{clusdiff} can generate a fascinating final dish from the recipe by fine-tuning the latent diffusion model. It was not until the past two years that few works began to explore the cooking process visualization \cite{cookingdiffusion, song2025makeanything, tip}. CookingDiffusion \cite{cookingdiffusion} first introduced this task by leveraging three innovative Memory Nets to model procedural prompts, which align with the corresponding step in a recipe, ensuring the semantic consistency of cooking. MakeAnything \cite{song2025makeanything} attempted to combine procedural images into a grid image and decomposed the static image into procedural sequences, enabling frame consistency for cooking process synthesis.

Despite the success of aforementioned studies, cooking process visualization still faces the following challenges. First, in the cooking process, the appearance of ingredients changes complexly and variously, making it difficult for models to generate the correct appearance in each frame. For example, in Figure \ref{fig1}, the chicken exhibits noticeable variation in some frames while retaining a consistent appearance in others.
Second, the generated cooking process is required to maintain frame coherency and a rational sequential order. However, most current models focus more on the semantic features of current step and ignore tracking the progress of cooking, leading to a chaotic sequential order or contradictory transitions between frames of the cooking process. 
Concretely, this omission of intermediate representation makes the model improperly integrate the semantics of previous step into the current step, causing frames to skip, reintroduce, or rearrange ingredients in an illogical manner. Without a mechanism for capturing how each step evolves from the previous one, the final generated sequence may stray from the normal cooking process.
As shown in Figure \ref{fig1}, in the early cooking process (Step 1-2), it is necessary to prepare ingredients such as scallion and pepper before proceeding with formal cooking (Step 3-4), while some previous methods may directly generate complete food images in the first few steps, causing an obvious chaotic order.

To address the above limitations, we propose a novel cooking process visualization approach named Chain-of-Cooking (CoCook). (1) For the first challenge, we discover that not all previous frames are useful for the current step. Meanwhile, along with the transformation and dissolution of ingredients, some visual information in historical frames becomes outdated. Therefore, we propose the Dynamic Patch Selection Module to dynamically select image patches from historically generated images that match the text of current step, which can provide a relevant visual reference for accurate generation of ingredient appearance, thereby enhancing the semantic consistency.
(2) For the second challenge, we design two innovative modules: the Semantic Evolution Module and a Bidirectional Chain-of-Thought (CoT) Guidance. Specifically, considering that each step is influenced by previous ones, we first initialize a latent prompt for the current step with the semantics of the previous cooking step. To better utilize the semantics of previous texts in each diffusion step, the Semantic Evolution module establishes semantical association between the current text and the latent prompts, then incorporates it into the latent features of diffusion model. Then the Bidirectional CoT Guidance updates the incorporated latent features through learnable latent prompts in both forward and backward diffusion processes. With the guidance of previous step information, the generated images can keep a rational sequential order and have better frame coherence across the cooking process.

In addition, most datasets for food image generation only have one image per recipe without intermediate images, which is unsuitable for cooking process visualization. Although few datasets such as RecipeQA \cite{recipeqa} contain food images for each cooking step, they have problems such as low image resolution and incomplete image-text pairs. To tackle this issue, we construct a dataset from cooking websites named CookViz, comprising high-quality sequential food images and complete intermediate texts for each cooking step, which provides comprehensive data support for studies of cooking process visualization.

Overall, our contributions can be summarized as follows:
\begin{itemize}
%\item We propose a novel method for cooking process visualization named Chain-of-Cooking, achieving frame coherent and semantically consistent cooking image generation.
\item We present a Dynamic Patch Selection Module, which retrieves the most relevant reference image patches to ensure the correct appearances of generated food image.

\item We introduce the Bidirectional Chain-of-Thought Guidance together with a Semantic Evolution Module. They enhance the rationality for coherent cooking process visualization.

\item We collect a dataset for cooking process visualization called CookViz, containing complete image-text pairs of the intermediate cooking process. 

\item Extensive qualitative and quantitative experiments demonstrate that our proposed model outperforms existing methods, especially in frame coherent and semantically consistent cooking process generation.

\end{itemize}

\section{Related Works}
\subsection{Food Image Generation}
Food image generation \cite{cookgan, cookgalip, clusdiff, foodfusion} has become a prominent research area, which is an intersection task of image generation and food analysis. Early studies such as ChefGAN \cite{chefgan} and CookGAN \cite{cookgan} focused on leveraging generative models to synthesize visually appealing food images from texts. Recent advancements focus on capturing fine-grained textures, ingredient interactions, and dataset qualities. For example, FoodFusion \cite{foodfusion} employs two distinct data cleaning methodologies to ensure the quality and accuracy of data and utilize LDM \cite{ldm} for realistic food image generations. ClusDiff \cite{clusdiff} generates images of multiple food categories through clustering methods. However, these food image generation works only generate the final dish images, but the object of our task is to generate images for each cooking step.

\begin{figure*}[t]
\centering
   \includegraphics[width=\textwidth]{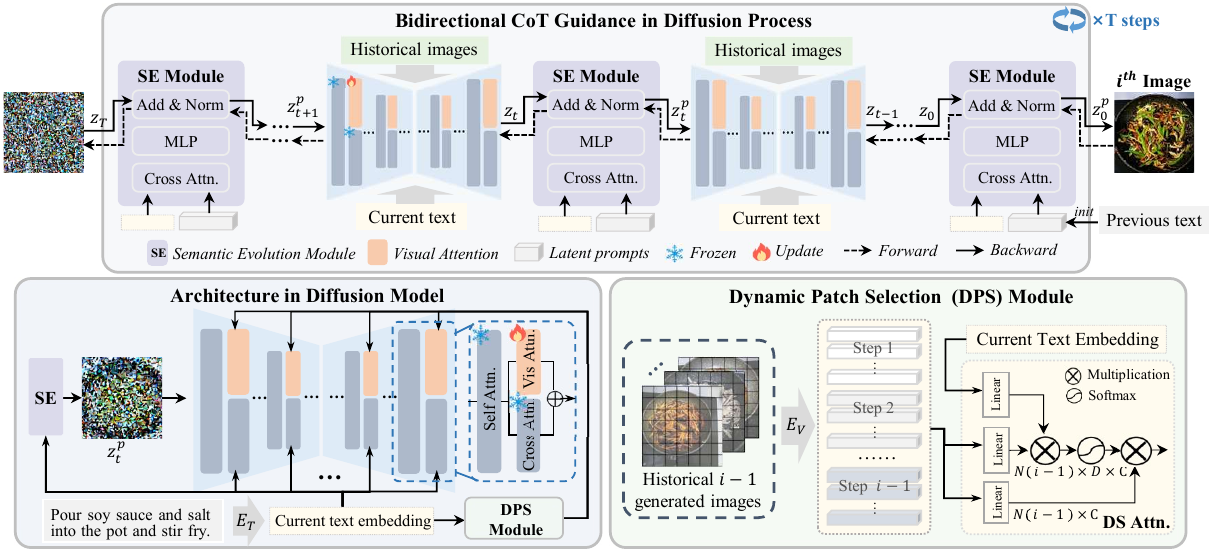}
   \caption{Overview of our proposed Chain-of-Cooking (CoCook) model, which consists of three specially designed methods: the Dynamic Patch Selection Module, the Semantic Evolution Module, and the Bidirectional Chain-of-Thought (CoT) Guidance.}
   \label{fig2}
\end{figure*}

\subsection{Sequential Image Generation}
Sequential image generation aims to generate a series of images that describe a story or a process. Story visualization is the typical task of sequential image generation. For example, StoryGen \cite{storygen} and Story-LDM \cite{storyldm} achieve coherent storyline generation by using contextual information in the story to generate the current frame. 
However, the story visualization task has fixed characters and simple scenes, while the combination and appearance of the ingredients change variously, and keeping the rational sequential order is essential. Therefore, the research of cooking process visualization is rare. MARS \cite{mars} explored the simultaneous generation of text and images by fine-tuning their pre-trained model with 10k unpublic recipes. CookingDiffusion \cite{cookingdiffusion} leverages three Memory Nets to model procedural prompts: text prompts that represent cooking steps, image prompts that correspond to cooking images, and multi-modal prompts that mix cooking steps and images, which ensure the consistent generation of cooking procedural images. TIP \cite{tip} leverages LLMs and multimodal generation models to generate procedural plans, but only focuses on the content of individual steps, resulting in inconsistency across the overall process.
Overall, attributed to the complexly varied ingredients at different steps and the lack of proper datasets, current works can only tackle sequential image generation of simple object and scene changes, making it difficult to visualize a coherent cooking process.

\subsection{Chain-of-Thought Reasoning}
Chain-of-Thought (CoT) reasoning \cite{wei2022chain, wang2022self, zhang2023multimodal} is an improved prompt technique that constructs intermediate steps to enhance the reasoning ability of model. It has been widely employed to solve complex reasoning problems, offering potential benefits including spatial layout planning ability, self-correction, and efficiency. 

Recently, CoT has been increasingly used in generation tasks. For text generation, DoT \cite{dot} adopts latent prompts to diffusion language models, which improve the reasoning ability of models and offer flexibility in trading-off computation for reasoning performance. CoT has also achieved great success in the image generation task, especially in spatial location reasoning and semantic consistency enhancement. For example, RPG \cite{rpg} introduces the extra MLLMs \cite{gpt4, team2023gemini} to generate rationales for region-wise planning, which makes up for the shortcomings of the generative model in terms of spatial reasoning ability. PARM \cite{parm} applies CoT to autoregressive image generation scenarios to explore whether they can reinforce image generation step-by-step. 
As a result, considering the complex multiple steps in the cooking process, we consider introducing CoT to our model for better reasoning ability. 
 
\section{Method}
\subsection{Problem Definition}
Cooking process visualization uses generative methods to synthesize an image sequence that is synchronized with the text and coherent with the previous images. Given a recipe with multiple cooking steps $\mathcal{S}=\{s_1, s_2, \cdots, s_n\}$, each step $s_i$ consists of a paired cooking text $\mathcal{T}_i$ and real image $\mathcal{V}_i$. The goal is to generate sequential cooking images that correspond to each input step $s_i$, which is formulated as follows:
\begin{equation}
    \hat{\mathcal{V}}_i = \Phi (\hat{\mathcal{V}}_i| \mathcal{T}_i, (\hat{\mathcal{V}}_{<i}, \mathcal{T}_{<i})) \ ,
\end{equation}
where $\Phi$ refers to the generative model, $\hat{\mathcal{V}}_i$ is the generated image at the $i^{th}$ cooking step. A cooking process model takes the current text $\mathcal{T}_i$, the previous text $\mathcal{T}_{i-1}$, and historical generated images $\hat{\mathcal{V}}_{<i}$ as condition, generating the image $\hat{\mathcal{V}}_i$ consistent with the current cooking text and coherent with previous frames.

\subsection{Model Overview}
As shown in Figure. \ref{fig2}, our proposed CoCook is built on the pre-trained Stable Diffusion model \cite{ldm}, and contains three specially designed modules: the Dynamic Patch Selection Module, the Semantic Evolution Module, and the Bidirectional CoT Guidance. Specifically, for the cooking step $i$, CoCook has four inputs: the current text embedding $e_i$ encoded by CLIP text encoder $E_T$ \cite{clip}, a noise vector $\epsilon$, the latent prompts $\mathcal{P}_t$ initialized from the previous text embedding $e_{i-1}$, and the historical generated images $\hat{\mathcal{V}}_{<i}$. We first initialize the latent prompt from $e_{i-1}$, then we build the semantical association with $e_i$ and $\mathcal{P}_t$, and integrate it with the diffusion latent features through the Semantic Evolution Module. During the training stage, the proposed Dynamic Patch Selection Module splits historically generated images into patches, which dynamically retrieves the most relevant patches as a reference for diffusion. We constantly update the latent prompts both in the forward and backward diffusion process through the Bidirectional CoT Guidance, and the parameters of diffusion model are frozen.

\subsection{Dynamic Patch Selection Module}
The Dynamic Patch Selection Module is designed to retrieve image patches from historically generated images that are most related to the textual content of current steps, and provides references for food visualization. This module consists of a historical image bank $B$, a CLIP image encoder $E_V$ \cite{vit-b/32}, and a Dynamic Selection Attention. The images generated for previous cooking steps are firstly saved in the historical image bank, which can be written as:
\begin{equation}
    B = \{ \hat{\mathcal{V}}_1, \hat{\mathcal{V}}_2, \ldots , \hat{\mathcal{V}}_{i-1} \} \ ,
\end{equation}

As depicted in the bottom right corner of Figure \ref{fig2}, each image in $B$ is split into $N$ patches and encoded by the image encoder $E_V$. Since some visual information in historical frames might be outdated, we introduce the Dynamic Selection Attention, which dynamically retrieves image patches that are most related to the textual content of current step. 

Given the textual description of the $i^{th}$ cooking step, the historical images patch features encoded by $E_V$ are denoted as $h_{< i}$, we respectively map the patch features $h_{< i}$ and the current text embedding $e_i$ through a Linear layer, and integrate $e_i$ with $h_{< i}$, which is formulated as:
\begin{equation}
    M = \mathrm{LN}(e_i) \otimes  \mathrm{LN}(h_{< i}) \ ,
\end{equation}
where $h_{<i}, M \in \mathbb{R}^{B\times N(i-1)\times D\times C}$. $B$ is the batch size, $N(i-1)$ is the patch numbers of images, $D$ is the dimension of text embedding $e_i$, and $C$ is the channel size. $\otimes$ represents the multiplication function. LN is the Linear layer.

The $M$ calculates the attention between each text token and image patches. To retrieve image patches based on the semantics of a whole sentence, we sum the weights of $M$ in the text dimension $D$ to obtain a new attention matrix. Then the $M$ and $h_{<i}$ are multiplied to obtain the filtered image patch features as follows:
\begin{equation}
   \tilde{M}_{b,n,c} = \mathrm{Softmax}(\sum_{d=1}^{D} M_{b,n,d,c}) \otimes \mathrm{LN}(h_{< i})\ ,
\end{equation}
where $\tilde{M}_{b,n,c} \in \mathbb{R}^{B\times N(i-1)\times C}$ represents the output feature of Dynamic Selection Attention. The $b,n, c$ represents the index of $B, N(i-1)$, and $C$ dimension respectively.

To utilize the selected patch features to guide image generation process, we copy the pre-trained Cross Attention as the Visual Attention, then use $\tilde{M}_{b,n,c}$ as the condition and fine-tune it during training. The output of Visual Attention is denoted as $F_{\mathcal{V}}$. We respectively map $F_{\mathcal{V}}$ and the output from frozen Cross Attention as query, key, and value. The feature after connection is denoted as:
\begin{equation}
    O =\mathrm{Softmax}(\frac{\mathbf{Q}(\mathbf{K})^{\top}}{\sqrt{d}})\mathbf{V}+\mathrm{Softmax}(\frac{\mathbf{Q}_{F_{\mathcal{V}}}(\mathbf{K}_{F_{\mathcal{V}}})^{\top}}{\sqrt{d}})\mathbf{V}_{F_{\mathcal{V}}} ,
\end{equation}
where $\mathbf{Q}, \mathbf{K}, \mathbf{V}$ represent the weight matrices of query, key, value from frozen Cross Attention. $\mathbf{Q}_{F_{\mathcal{V}}}, \mathbf{K}_{F_{\mathcal{V}}}, \mathbf{V}_{F_{\mathcal{V}}}$ represent the weight matrices of query, key, and value from Visual Attention. $O$ is the final output feature after concatenation.

\subsection{Semantic Evolution Module}
To ensure the rational sequential order for each step of image generation, we integrate the semantics of the previous step with current text to provide guidance and constraints in maintaining correct order across the cooking process. To achieve this, we propose the Semantic Evolution Module to explore the semantical association between the current step and the previous step, avoiding irrelevant information in the previous step that misleads the generation.

Specifically, for the diffusion step $t$ in Figure \ref{fig2}, the proposed Semantic Evolution Module has three inputs: the latent features $z_{t}$, the current encoded text embedding $e_i$, and latent prompts $\mathcal{P}_t$ that initialized from the previous text $\mathcal{T}_{i-1}$. $\mathcal{T}_{i-1}$ is first encoded by the CLIP text encoder and mapped into latent prompts through a linear layer. Then based on $e_i$ and $\mathcal{P}_t$, we employ a learnable cross attention to obtain attention features $z_t'$ aligned to $\mathcal{P}_t$, which can provide the semantic cues for the current step. Specifically, we regard the $\mathcal{P}_t$ as query, the $e_i$ as key and value respectively, which can be formulated as follows:
\begin{equation}
    z_t' = \mathrm{CrossAttn}(\mathbf{Q}={\mathcal{P}_t},\mathbf{K}={e_i},\mathbf{V}={e_i}) \ ,
\end{equation}

After that, by incorporating $z_t'$ with latent features of current step of generation $z_{t}$ through adding and normalization operation, we have the output of our Semantic Evolution Module as follows:
\begin{equation}
    \mathbf{z}_t^p=\mathrm{AddNorm}\left(\mathbf{z}_{t},\mathrm{MLP}(\mathbf{z}_t^{\prime})\right) \ ,
\end{equation}
where $\mathrm{MLP(\cdot)}$ denotes the MLP layer used to transform the dimension of $z_t'$. $\mathbf{z}_t^p$ represents the output of our Semantic Evolution Module at the current cooking step, embodying both historical and current semantics. By progressively updating semantic cues in the diffusion process, it guarantees a coherent flow of semantics across the cooking process during generation.

\subsection{Bidirectional Chain-of-Thought Guidance}
The forward and backward process of existing diffusion models often lack additional constraints or guidance that would enhance reasoning during generation. Inspired by the research of DoT \cite{dot}, we propose the Bidirectional CoT guidance to further enhance the reasoning ability of diffusion model. As shown in the top part of Figure \ref{fig2}, in the forward process, $z_{t-1}^p$ provides a semantic condition that includes the previous step to generate image $z_t$ of current cooking step. In the backward process, $z_{t+1}^p$ can guide the diffusion model to reconstruct the current image $z_t$. Through this bidirectional guidance in diffusion process, the frame coherence and rationality of cooking process visualization are better guaranteed.

Consequently, given a set of training data for cooking process visualization, we have two training objectives for the CoCook model: one is fine-tuning from the pre-trained diffusion model, while the other is focused on the semantic evolution associated with the cooking steps. To be specific, fine-tuning the pre-trained diffusion model aims to generate images by incorporating a noise map $\epsilon \in \mathcal{N}(0,1)$, the conditional textual condition $c_t^{\mathcal{T}}$ that encoded by CLIP text encoder, and visual condition $c_t^{\mathcal{V}}$ that derived from historical image patches. Given the $z_t^{p}$ that guided by latent prompts, and conditional embeddings $c_t^{\mathcal{V}}$, $c_t^{\mathcal{T}}$, $\mathcal{\epsilon_{\theta}}$ is utilized to predict the noise $\epsilon$ at the time step $t$. The loss function $\mathcal{L}_{CE}$ is written as:

\begin{equation}
    \mathcal{L}_{CE} = \mathbb{E}_{z_t^{p},c_t^{\mathcal{V}},c_t^{\mathcal{T}},\epsilon\sim\mathcal{N}(0,1),t}[\parallel\epsilon-\epsilon_{\theta}(z_t^{p}|c_t^{\mathcal{V}},c_t^{\mathcal{T}},t)\parallel_{2}^{2}] \ ,
\end{equation}
where $\epsilon_{\theta}$ is the denoiser that is used to denoise the image in the diffusion latent space, and $t=0,1, \cdots , T$.

Furthermore, in the diffusion process to generate images, latent prompts assist the images in keeping a rational sequential order based on the semantics of the previous and current cooking step. Therefore, we introduce a semantically consistent loss $\mathcal{L}_{CS}$ to constrain latent prompts $\mathcal{P}_t$ and cooking instructions $\mathcal{T}_i$, making them closer to the cooking step in the semantic space.
\begin{equation}
    \mathcal{L}_{CS}(\mathcal{P}_t,\mathcal{T}_i) = 1-\frac{\mathcal{P}_{t}^{\top}\mathcal{T}_i}{\parallel\mathcal{P}_{t}^{\top}\parallel\parallel \mathcal{T}_i\parallel} \ ,
\end{equation}

Minimizing $\mathcal{L}_{CS}$ brings the latent prompts closer to the current cooking step in a shared semantic space, and preserves a correct sequential semantic order. The final loss function $\mathcal{L}$ is written as:
\begin{equation}
    \mathcal{L} = \mathcal{L}_{CE} + \eta \mathcal{L}_{CS} \ ,
\end{equation}
where  $\mathcal{L}_{CE}$ is the cross-entropy loss and $\mathcal{L}_{CS}$ minimizes the semantic distance between the latent prompts and the cooking instruction. $\eta$ is the loss weight to adjust the influence of different losses.

For the predicted noise, we employ the classifier-free guidance in inference and the calculation can be expressed as:
\begin{equation}
    \hat\epsilon_{\theta} (z_t^{p}|c_t^{\mathcal{V}},c_t^{\mathcal{T}} ,t) =(w+1)\epsilon_{\theta}(z_t^{p}|c_t^{\mathcal{V}},c_t^{\mathcal{T}} ,t)-w\epsilon_{\theta}(z_t^{p}|t) \ ,
\end{equation}
where $w$ is the guidance scale. 

During the inference stage, we generate image at the first step directly from current cooking instruction without the guidance from historical images. When the cooking step $i>1$, we select image patches from previously synthesized frames along with the cooking instructions, which are treated as conditions to synthesize the image sequence in an auto-regressive manner. 

\section{Dataset}
Obtaining the complete image-text pairs of a whole cooking process is challenging and most publicly available datasets have poor image resolution or unpaired images and texts of the cooking process. Thereby, we construct a high-quality dataset for cooking process visualization, termed as CookViz. 

Specifically, we download a large number of cooking processes from popular cooking websites, which contain step-by-step cooking texts and high-quality intermediate images. To ensure the complexity of datasets, each recipe was required to have at least three cooking steps. After that, we use regular expressions to filter out irrelevant characters such as emoji in the text. Moreover, to eliminate the impact of image watermark on the generation model, we use the OCR \cite{liu2022convnet} model to locate the textual watermark, and then utilize the SD-Inpainting model \cite{ldm} to remove watermarks in images.

\begin{table}[t]
\centering
\caption{Comparison of existing datasets that are suitable for the cooking process visualization task.}
\begin{tabular}{l|c|c|c}
\toprule
\textbf{Dataset} & \textbf{Public} & \textbf{\#Frames} & \textbf{Language} \\
\midrule
 Cookpad \cite{cookpad}  & False & 1,642,450 & Japanese  \\
 MIAIS \cite{miais} & False & 121,547 & Japanese \\
 MM-Res \cite{mm-res}   & False     & 227,082 & English\\
 RecipeQA\cite{recipeqa}     & True      & approx. 20,000 & English \\
\textbf{CookViz} & True & 40,362 & English\\
\bottomrule
\end{tabular}
%\vspace{-1.2em}
\label{table1}
\end{table}

\begin{figure*}[htbp]
\centering
   \includegraphics[width=\textwidth]{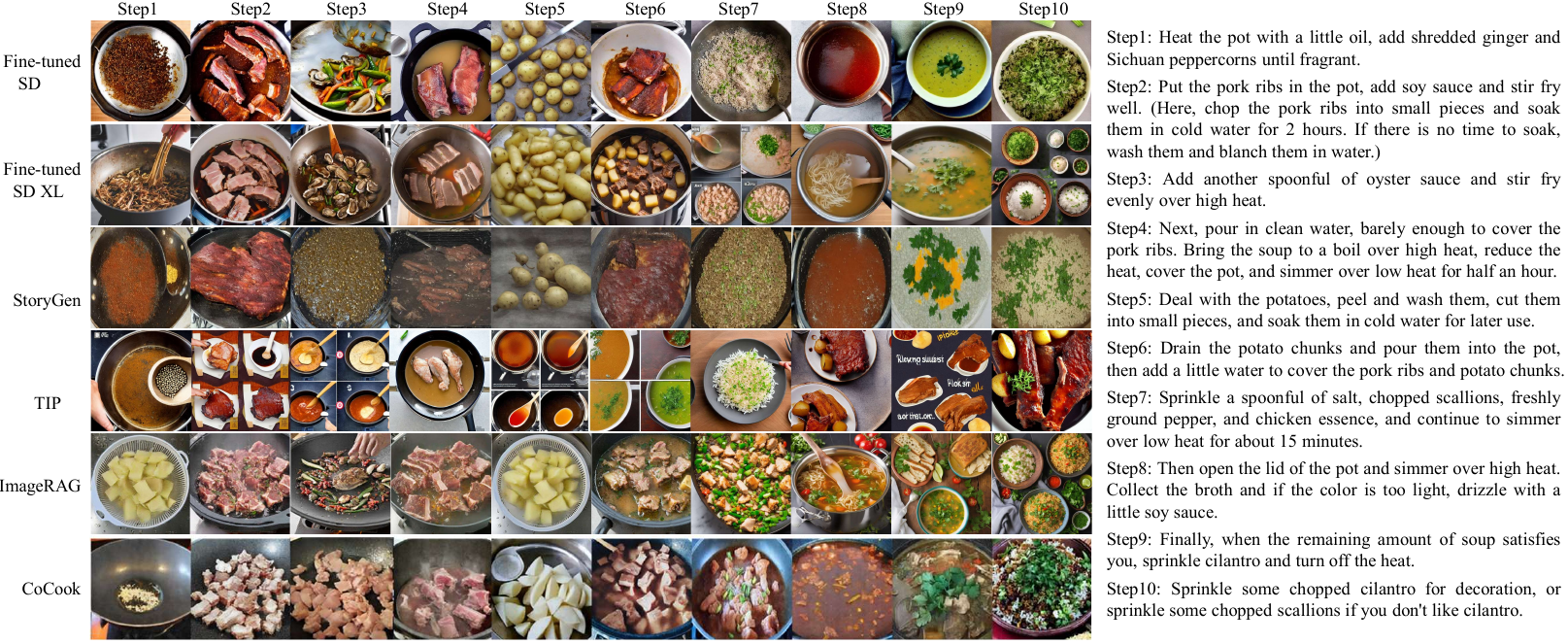}
   \caption{Visualization results of generated procedural cooking images compared with the fine-tuned stable diffusion models\cite{ldm, sdxl}, StoryGen\cite{storygen}, TIP\cite{tip}, and our CoCook model.}
   \label{fig3}
\end{figure*}

We compare our proposed CookViz dataset with existing datasets in Table \ref{table1}. Although there are already some datasets for the cooking process visualization task, they are not suitable for our task due to language differences or unavailability. Compared to the only publicly available dataset RecipeQA, our dataset has about 40k cooking steps with higher image resolution and cleaner texts, showing a significant influence on cooking process visualization. 

To train and evaluate the models in experiments, we divide the training set and testing set of our CookViz dataset with the ratio 9:1. In addition, to comprehensively evaluate the generation capability of our model, we experiment on the public dataset RecipeQA, which consists of approximately 20K recipes from 22 food categories.

\section{Experiments}
In this section, we outline our experimental setup first, then we introduce the evaluation metrics, and also conduct human evaluations to assess the text-image consistency and frame coherency. Finally, we present comparisons against the state-of-the-art methods and include comprehensive ablation studies to demonstrate the effectiveness of each proposed module in our framework. We compare the results synthesized by our proposed CoCook and those generated by TIP \cite{tip}, StoryGen \cite{storygen}, Stable Diffusion v1.5 \cite{ldm}, Stable Diffusion XL \cite{sdxl}, and the retrieval-based model ImageRAG \cite{shalev2025imagerag}. To fairly assess the performance of compared models, we fine-tune the Stable Diffusion v1.5 and XL models with LoRA \cite{lora} on RecipeQA \cite{recipeqa} and CookViz datasets. 

\subsection{Experimental Setup}
Our model is built on the pre-trained stable diffusion v1.5 model \cite{ldm}. We train our model using an entire cooking process as one batch with a learning rate of $1\times 10^{-5}$, which includes several intermediate image-text pairs. We conduct experiments on 4 NVIDIA RTX 3090 with 24G memory, employing half-precision training. The input images of training are resized to $512 \times 512$, and the feature dimension is set to 768. The hyperparameter $\eta$ in loss function is set to 0.1. During inference, we adopt DDIM scheduler with 50 steps of sampling, and the guidance weight $w$ is 7.5.

\subsection{Evaluation Metrics}
We adopt three widely used evaluation metrics to evaluate our generated images of the cooking process, including the FID \cite{fid}, CLIP-T \cite{clip}, CLIP-I \cite{clip} and DreamSim \cite{fu2023dreamsim}.

\noindent \textbf{Fréchet Inception Distance (FID):} We use FID to measure the realism of the generated cooking procedural images. FID compares the distribution of real images to generated images in the feature space of a pre-trained Inception network \cite{inceptionv3}. A lower FID value indicates that the synthesized images have better quality.

\noindent \textbf{CLIP-I:} Inspired by the video frame consistency in video generation \cite{wang2023dreamvideo}, we leverage a pretrained CLIP model to calculate the similarity of adjacent generated frames and take the average value. A higher CLIP-I score implies that the generated images are more coherent.

\noindent \textbf{CLIP-T:} To evaluate how well the generated sequential image covers the complex cooking instructions, we utilize CLIP-T to measure text-image similarity scores between the cooking instructions and the corresponding procedural images. A higher CLIP-T score means the better semantic consistency between images and texts.

\noindent \textbf{DreamSim (DSim):} DreamSim is used to assess the coherence of image sequences. It is trained by concatenating CLIP, OpenCLIP \cite{openclip}, and DINO \cite{dino} embeddings, and then finetuning on human perceptual judgements, achieving better alignment with human similarity judgments than existing metrics.

\subsection{Qualitative Analysis}
We first demonstrate the generated images between our CoCook and existing methods in terms of visualization quality during the cooking process.
The results in Figure \ref{fig3} demonstrate that the fine-tuned SD v1.5 and XL models generate rough semantics of the input cooking step, and fail to perform well in terms of frame coherency and semantic accuracy. For example, the potatoes at step 5 should be cut into small pieces, while among all the comparison methods, our model can correctly generate potato pieces. Furthermore, we find that although StoryGen is superior to Stable Diffusion models in terms of frame coherency, it seems to have poor image quality, which is reflected in the shape of ingredients like pork ribs. The images generated by TIP model, which use LLMs to generate the visual imaginative prompt for cooking process, perform unwell in terms of food appearance and semantic consistency. The spatial layout of the images is also unreasonable, for example, multiple sub-scenes may appear simultaneously in step 2 and step 3 of the TIP model. ImageRAG retrieves the entire image as a condition, resulting in some frames are overly consistent.
In contrast, our proposed CoCook visually outperforms all the above methods, which indicates that CoCook achieves the best balance between frame coherency and semantic accuracy.

\subsection{Quantitative Analysis}

In this section, we measure the quantitative results of generated procedural images in Table \ref{table2}. To assess the quality of ground truth (GT), we also provide CLIP-T and CLIP-I of the RecipeQA and CookViz datasets. It should be noted that the calculation of FID requires both real images and generated images, so the FID metric of ground truth is empty. The value of CLIP-I is 1, indicating that the real images in the dataset are coherent by default. 

\begin{table*}[t]
%\hspace{0.001\textwidth}
\centering
\setlength{\tabcolsep}{0.8mm}
%\small
\caption{Quantitative evaluations and human evaluations (Consis., Coher.) on the RecipeQA and CookViz datasets.}
\resizebox{\linewidth}{!}{
\begin{tabular}{l|c|c|c|c|c|c|c|c|c|c|c|c}
\toprule
\multirow{2}{*}{\textbf{Model}} & \multicolumn{6}{c|}{\textbf{RecipeQA}} & \multicolumn{6}{c}{\textbf{CookViz}} \\ \cmidrule(r){2-13}
& \textbf{FID↓} & \textbf{CLIP-T↑} & \textbf{CLIP-I↑} & \textbf{DSim↓} & \textbf{Consis.↑} & \textbf{Coher.↑} & \textbf{FID↓} & \textbf{CLIP-T↑} & \textbf{CLIP-I↑} & \textbf{DSim↓} & \textbf{Consis.↑} & \textbf{Coher.↑} \\
\midrule
 GT                                 & -     & 0.207 & 1     & 1 & 4.12 & 3.98  & -     & 0.215 & 1     & 1  & 4.76  & 4.80 \\
 fine-tuned SD \cite{ldm}           & 18.35 & 0.200 & 0.733 & 0.6145 & 3.70 & 3.44  & 13.80 & 0.210 & 0.813 & 0.6150 & 3.90 & 3.84 \\
 fine-tuned SDXL \cite{sdxl}        & 18.76 & 0.202 & 0.730 & 0.6308 & 3.78 & 3.46  & 13.81 & 0.212 & 0.813 & 0.6163 & 3.96 & 3.84 \\
 StoryGen \cite{storygen}           & 17.36 & 0.204 & 0.737 & 0.6105 & 3.62 & 3.62  & 16.77 & 0.213 & 0.818 & 0.6065 & 3.84 & 4.02 \\
 TIP  \cite{tip}                    & 18.69 & 0.208 & 0.734 & 0.62440& 3.60 & 3.38  & 14.65 & 0.215 & 0.814 & 0.6127 & 3.80 & 3.96 \\
 ImageRAG \cite{shalev2025imagerag} & 16.81 & 0.204 & 0.735 & 0.6120 & 3.72 & 3.70  & 13.91 & 0.212 &0.815  & 0.6100 & 4.04 & 3.98 \\
\textbf{CoCook}   & \textbf{12.43} & \textbf{0.209} & \textbf{0.739} & \textbf{0.5941} & \textbf{3.84} & \textbf{3.86} & \textbf{12.28} & \textbf{0.218} & \textbf{0.820} & \textbf{0.5851} & \textbf{4.22} & \textbf{4.30} \\
\bottomrule
\end{tabular}}
\label{table2}
\end{table*}

The fine-tuned SD v1.5 and SD XL methods maintain image quality but struggle to realize coherency in generated procedural images. As a comparison, our model demonstrates a significant improvement compared to Stable Diffusion models, especially in terms of frame coherency and semantic alignment, indicating that it can generate images that are highly consistent with the given cooking instructions and previous steps. Compared with StoryGen and ImageRAG, which can generate coherent procedural images, it is still inferior to our method in terms of semantic consistency and image quality during the cooking process. Moreover, TIP demonstrates moderate performance surpassing other methods in CLIP-T while still trailing behind CoCook. Because TIP optimizes text by introducing additional language models, it eliminates irrelevant words in the text, especially on the RecipeQA dataset.

Besides the comparison between different models, it can also be noticed that the CLIP-T of our CoCook dataset is higher than the ground truth. This is because the real cooking images often display unexpected objects or details that are not explicitly mentioned in the recipe text (such as extra utensils, cluttered backgrounds), and the generated images will focus purely on elements related to the recipe. Thereby, the CLIP-T may be higher than ground truth, this phenomenon has also been observed in the previous work \cite{storygen}. 
%Moreover, the CLIP-T of GT on CookViz outperforms the RecipeQA dataset, which reflects in the higher image resolution and absence of redundant text.

\vspace{-3pt}
\subsection{Human evaluations}
Considering that the above metrics may not accurately assess the generated images, and there are no standardized metrics to evaluate the consistency of procedural images within the cooking process, we further conduct human evaluations for the comprehensive comparison of image-text consistency and frame coherency. 

Referring to similar studies for procedural image generation\cite{storygen}, we randomly select an equal number of generated samples of our CoCook and other compared works. We provide guidelines for participants that involve two aspects: (1) the semantic consistency between generated images and cooking instructions; (2) the coherency of adjacent generated images. Concretely, the consistency refers to whether the ingredients with corresponding shapes and changes in the generated image are consistent with text inputs. The coherency refers to the adjacent frames need to keep a sequential order and cannot undergo changes that violate common sense. Each time we randomly sample a cooking process from these selected samples, and 50 participants are invited to rate them with a score ranging from 1 to 5. Higher scores indicate better samples. 

As illustrated in Table \ref{table2}, our proposed CoCook achieves competitive performance in both consistency and coherency compared with existing methods. Although ground truth remains the upper bound, the balance between consistency and coherence shows the effectiveness of our approach in aligning textual instructions with coherent visual features across different datasets.

\begin{figure*}[t]
\centering
   \includegraphics[width=\textwidth]{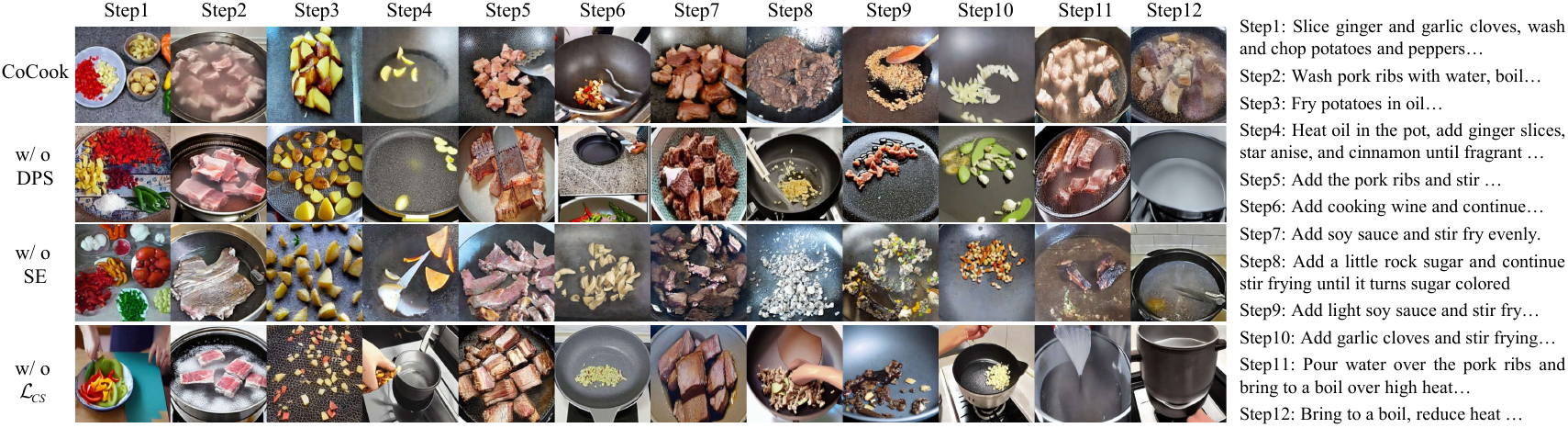}
   \caption{Ablation experiments of key components in CoCook model, including removing the Dynamic Patch Selection Module (DPS), the Semantic Evolution Module (SE), and the semantic consistency loss $\mathcal{L}_{CS}$.}
   \label{fig4}
\end{figure*}

\vspace{-3pt}
\subsection{Ablation Studies}
To demonstrate the effectiveness of our proposed module, we conduct ablation studies using quantitative metrics and qualitative visualization. We ablate our method by removing key components: the Dynamic Patch Selection Module (DPS), the Semantic Evolution Module (SE), and the semantically consistent loss $\mathcal{L}_{CS}$. We also compare methods of historical visual feature selection, including frame selection and patch selection.

\noindent \textbf{Effect of key components.} As shown in Table \ref{table4}, removing the DPS Module leads to notable drops in both FID and CLIP-I. This demonstrates that the DPS Module successfully ensures visual fidelity and refines the frame coherency of generated cooking images. Moreover, results of removing the SE Module and semantically consistent loss $\mathcal{L}_{CS}$ mainly affect the CLIP-T and CLIP-I, which shows that these two components contribute crucial step-by-step reasoning that maintains the rational sequential order and image coherency across the cooking process.

\begin{table}[t]
\centering
\setlength{\tabcolsep}{0.7mm}
\caption{Ablation experiments of key components and image selection methods on the RecipeQA and CookViz datasets.}
\begin{tabular}{l|c|c|c|c|c|c}
\toprule
\multirow{2}{*}{\textbf{Model}} & \multicolumn{3}{c|}{\textbf{RecipeQA}} & \multicolumn{3}{c}{\textbf{CookViz}} \\ \cmidrule(r){2-7}
 & \textbf{FID↓} & \textbf{CLIP-T↑} & \textbf{CLIP-I↑} & \textbf{FID↓} & \textbf{CLIP-T↑} & \textbf{CLIP-I↑} \\
\midrule
 CoCook           & 12.43 & 0.209 & 0.739 & 12.28 & 0.218 & 0.820 \\
 \ \ w/ o DPS   & 14.05 & 0.207 & 0.732 & 13.76 & 0.217  & 0.817 \\
 \ \ w/ o SE  & 12.92 & 0.204 & 0.736 & 12.89 & 0.215  & 0.819 \\
 \ \ w/ o $\mathcal{L}_{CS}$  & 12.60 & 0.205 & 0.735 & 12.41 & 0.215  & 0.817 \\
 \midrule
 \ \ w/ frame        & 12.84 & 0.204 & 0.732 & 12.72 & 0.216 & 0.820 \\
 \ \ w/ patch        & 12.43 & 0.209 & 0.739 & 12.28 & 0.218 & 0.820 \\
\bottomrule
\end{tabular}
\label{table4}
\end{table}

\noindent \textbf{Effect of visual reference selection.} To explore the influence of different selection methods for historical visual references on the generated results, we conducted experiments on selecting the entire images (w/ frame) and image patches (w/ patch) separately. The ``w/ frame'' setting provides moderate performance and enhances the iterative selection of key frames, helping to maintain the image coherency of cooking process. However, it still lags behind ``w/ patch'', indicating that focusing on local features, especially dynamic patch selection, can better capture subtle changes and eliminate interference from irrelevant image contents, thereby improving the quality of generated images.

\noindent \textbf{Qualitative Visualization.} In Figure \ref{fig4}, we notice the inconsistencies in both ingredient appearance and cooking utensils across consecutive steps without DPS. Omitting the Semantic Evolution Module results in a misalignment between recipe instructions and image details. While the cooking steps still unfold sequentially, the overall appearance of ingredients (like the color and texture of the pork in Steps 5–8) diverges from the intended process. This suggests the absence of compromises the model’s ability to maintain coherent, contextual transitions. Moreover, removing the semantic consistency loss $\mathcal{L}_{CS}$ weakens the link between texts and evolving visual features. Although some rough ingredient progression is retained, such as the addition of garlic in Step 1 or star anise in Step 4 is not consistently reflected in subsequent frames. Consequently, the logic of cooking process is often muddled. By contrast, our CoCook model effectively balances these visual and semantic constraints: each frame accurately reflects the relevant instructions (e.g., frying and color changes), while ingredients transform consistently across the 12 steps. This comparison illustrates how each proposed component contributes to achieving a coherent and logical cooking visualization sequence.

\begin{figure}[t]
    \centering
    \includegraphics[width=\columnwidth]{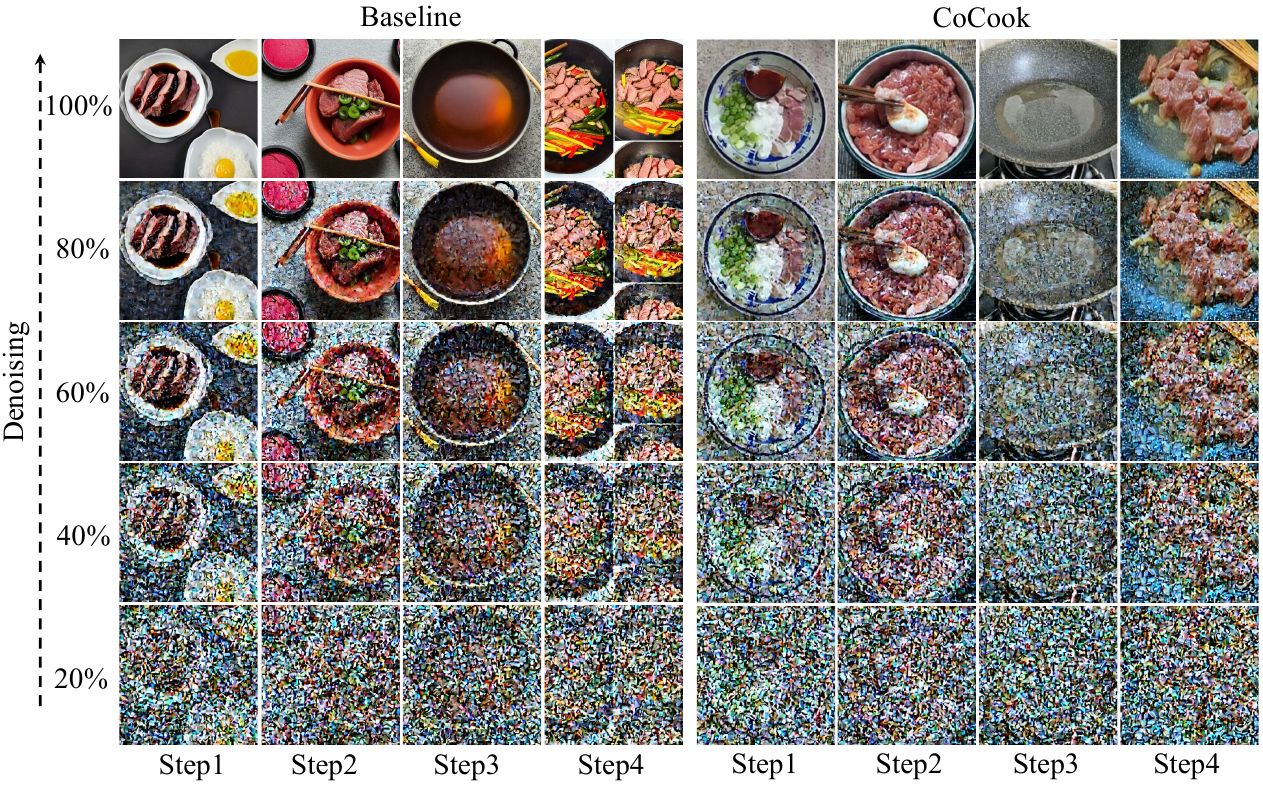}
    \caption{The intermediate decoded images of denoising process, including Baseline (SD v1.5) and CoCook. The cooking steps including Step 1: Prepare sliced meat, scallions, garlic, salt, starch, and soy sauce; Step 2: Add salt, soy sauce, and starch, marinate the meat slices for 20 minutes; Step 3: Pour oil into the pot and heat it up; Step 4: Pour the marinated meat slices into the pot and stir fry evenly.}
    \label{fig5}
\end{figure}

The intermediate decoded images of the denoising process in Figure \ref{fig5} indicate that our CoCook presents more reasonable contours in the early denoising stage. For example, in the denoising 20\% and 40\% stages, although the overall image still has strong noise, the ingredient contours of CoCook like meat slices and scallions show consistency with the cooking recipes. In the denoising 60-100\%, the images generated by CoCook can clearly recognize different cooking scenes and visually synchronize with the text descriptions, while the results of Baseline often lack coherence between different steps, such as the changes in dish or cooking utensils, making it difficult to reflect a logical cooking process. And the Baseline generates a final dish at Step 1 and exists an chaotic order.

\vspace{-2pt}
\section{Limitations and Future Work}
Despite the notable progress achieved by our proposed Chain-of-Cooking, there still remain several limitations. First, our model mainly focuses on the changes in cooking steps, and is still not robust enough for complex scenarios such as frequent background changes or multi-person collaborations. Furthermore, during the cooking process, ingredients often undergo dynamic changes such as multiple cuts, kneading, and heating. The current model relies on cooking texts and partial historical image information for inference. For dishes with extremely high detail requirements, the generated details still have deviations. 

In our future work, we can explore enriching the contextual understanding of model by incorporating scene parsing and dynamic background representations, which will help the model adapt to less controlled or more collaborative cooking environments. Furthermore, we consider integrating fine-grained temporal modeling, leveraging advanced generative modules or spatiotemporal attention mechanisms to capture nuanced ingredient changes.

\vspace{-2pt}
\section{Conclusion}
In this paper, we introduce Chain-of-Cooking, a novel method that can generate procedural images for the cooking process visualization task. By introducing the Dynamic Patch Selection Module and Bidirectional Chain-of-Thought Guidance, our method generates semantically consistent and visually coherent food images that reflect each intermediate step in a recipe. Moreover, we propose CookViz, a new dataset with complete intermediate images and aligned texts for the cooking process, enabling comprehensive research on cooking process visualization and broadening the application scenarios of food image generation. Extensive experiments on both CookViz and the public RecipeQA dataset demonstrate that our proposed method achieves notable improvements in FID, CLIP-based metrics, and human evaluations over existing works. CoCook successfully balances semantic accuracy with frame coherency, even in scenarios with sequential order and diverse ingredient transformations.

\begin{acks}
This work was supported by the National Natural Science Foundation of China under Grants No.62325206, the Key Research and Development Program of Jiangsu Province under Grant BE2023016-4, and the Postgraduate Research \& Practice Innovation Program of Jiangsu Province under Grant KYCX23\_1033.
\end{acks}
\vspace{-2pt}
\bibliographystyle{ACM-Reference-Format}
\bibliography{sample-base}

\end{document}